\documentclass[]{article}
\usepackage[margin=1in]{geometry}
\usepackage{longtable}
\usepackage{url}  
\usepackage{booktabs}      
\usepackage{amsfonts}       
\usepackage{nicefrac}       
\usepackage{microtype}      
\usepackage{xcolor}         
\usepackage{amsmath}
\usepackage{mathtools}
\usepackage{amsthm}
\usepackage{amsmath}
\usepackage{amssymb}
\usepackage{bbm}
\usepackage{tikz} 
\usepackage{wrapfig}
\usepackage{graphicx}
\usepackage{floatrow}
\usepackage{enumitem}
\usepackage[suppress]{color-edits}
\usepackage{caption}
\usepackage{subcaption}
\usepackage{authblk}
\usepackage{multirow}
\usepackage[colorlinks,allcolors=purple]{hyperref}
\PassOptionsToPackage{sort&compress}{natbib}
\usepackage[round, sort]{natbib}
\usepackage{booktabs} 
\usepackage[ruled]{algorithm2e} 
\usepackage{tikz}
\usetikzlibrary{arrows}
\usetikzlibrary{shapes.multipart}
\usepackage{caption}
\usepackage[all]{nowidow}
\usepackage[font=small,labelfont=bf]{caption}
\usepackage[utf8]{inputenc} 
\usepackage[T1]{fontenc}    
\usepackage[capitalise,nameinlink]{cleveref}
\usepackage{mdframed}
\usepackage{color,soul}

\newmdtheoremenv{theo}{Theorem}

\SetAlFnt{\small}
\SetAlCapFnt{\small}
\SetAlCapNameFnt{\small}
\SetAlCapHSkip{0pt}
\IncMargin{-\parindent}
\usepackage{color-edits}
\addauthor{kc}{blue}
\addauthor{lw}{purple}
\addauthor{aj}{green}
\addauthor{aw}{red}
\addauthor{mj}{magenta}

\usepackage[most]{tcolorbox}

\newtcolorbox{tabframebox}[1][]{breakable,sharp corners,boxrule=0.5pt,colback=white,halign=center,#1}

\title{Counterfactual Explanations and Predictive Models to Enhance Clinical Decision-Making in Schizophrenia using Digital Phenotyping}

\author[1]{Juan Sebastián Cañas}
\author[1]{Francisco Gómez}
\author[2]{Omar Costilla-Reyes\footnote{corresponding author: costilla@mit.edu. Working paper.}}
\affil[1]{Universidad Nacional de Colombia}
\affil[2]{CSAIL MIT}

\begin{document}
\maketitle

\begin{abstract}
\noindent Clinical practice in psychiatry is burdened with the increased demand for healthcare services and the scarce resources available. New paradigms of health data powered with machine learning techniques could open the possibility to improve clinical workflow in critical stages of clinical assessment and treatment in psychiatry. 
In this work, we propose a machine learning system capable of predicting, detecting, and explaining individual changes in symptoms of patients with Schizophrenia by using behavioral digital phenotyping data. We forecast symptoms of patients with an error rate below $10\%$. 
The system detects decreases in symptoms using changepoint algorithms and uses counterfactual explanations as a recourse in a simulated continuous monitoring scenario in healthcare.  Overall, this study offers valuable insights into the performance and potential of counterfactual explanations, predictive models, and change-point detection within a simulated clinical workflow. These findings lay the foundation for further research to explore additional facets of the workflow, aiming to enhance its effectiveness and applicability in real-world healthcare settings. By leveraging these components, the goal is to develop an actionable, interpretable, and trustworthy integrative decision support system that combines real-time clinical assessments with sensor-based inputs.

\end{abstract}




\maketitle
\pagestyle{plain} 
\section{Introduction}
Schizophrenia spectrum disorders (SSDs) are a family of serious mental illnesses (SMI) affecting approximately 24 million people worldwide. There are still challenges in providing healthcare access and treatment. People with SSDs that do not receive specialist mental health care are close to $70\%$ and just $30\%$ of them fully recover \cite{WHOschizophrenia}. Current clinical practices, such as conventional face-to-face assessments, are inefficient in detecting early observable behavioral precursors of schizophrenia, cannot scale and are not optimal to detect dynamic behavioral changes, as the nature of such diseases. Consequently, this leads to intervention at late stages of relevant clinical events \cite{Ben-Zeev2017}. 

Mobile computing technology has opened the possibility for the study of behavoral health conditions in a new way. The act of measurement no longer needs to be confined to clinics or research laboratories. Instead, it can be carried out in real-world settings. As a consequence, a new source of clinical data can be made available to turn it into biomedical knowledge and clinical insights \cite{sim2019mobile}. Given that these digital fingerprints reflect lived experiences of people in their natural environments, with the granular temporal resolution, it might be possible to leverage them to develop precise and temporally dynamic digital disease phenotypes and markers to diagnose and treat psychiatric illnesses and others \cite{Perez-Pozuelo2021}. In this research, we explore the area of \textit{digital phenotyping}, understood as the in-situ quantification of the individual-level human phenotype using data from digital devices \cite{Onnela2016,Mohr2017} for the study of schizophrenia.

In conventional clinical practice, healthcare providers engage in a process of gathering patient information, employing their clinical expertise to make informed decisions, and subsequently documenting their findings. In contrast, \textit{integrative decision support systems} (IDSS), can actively request clinically relevant information or gather the data, show the results to clinicians, and support decisions that clinicians still need to make \cite{yu2018artificial}. The goal of this research is to create actionable clinical systems capable of exploiting newly available data from digital phenotyping empowered with data science tools that could have an impact on the clinical task, such as early detection and intervention in mental healthcare \cite{russ2019data}. 

Driven by the challenge of intervening in the advanced stages of SSDs and recognizing the potential benefits of mobile computing in clinical practice, we have formulated the following research question: \textit{How can we effectively characterize the transitions between different symptom states for patients diagnosed with Schizophrenia using Digital Phenotyping?}

\begin{figure}
  \includegraphics[scale=0.7]{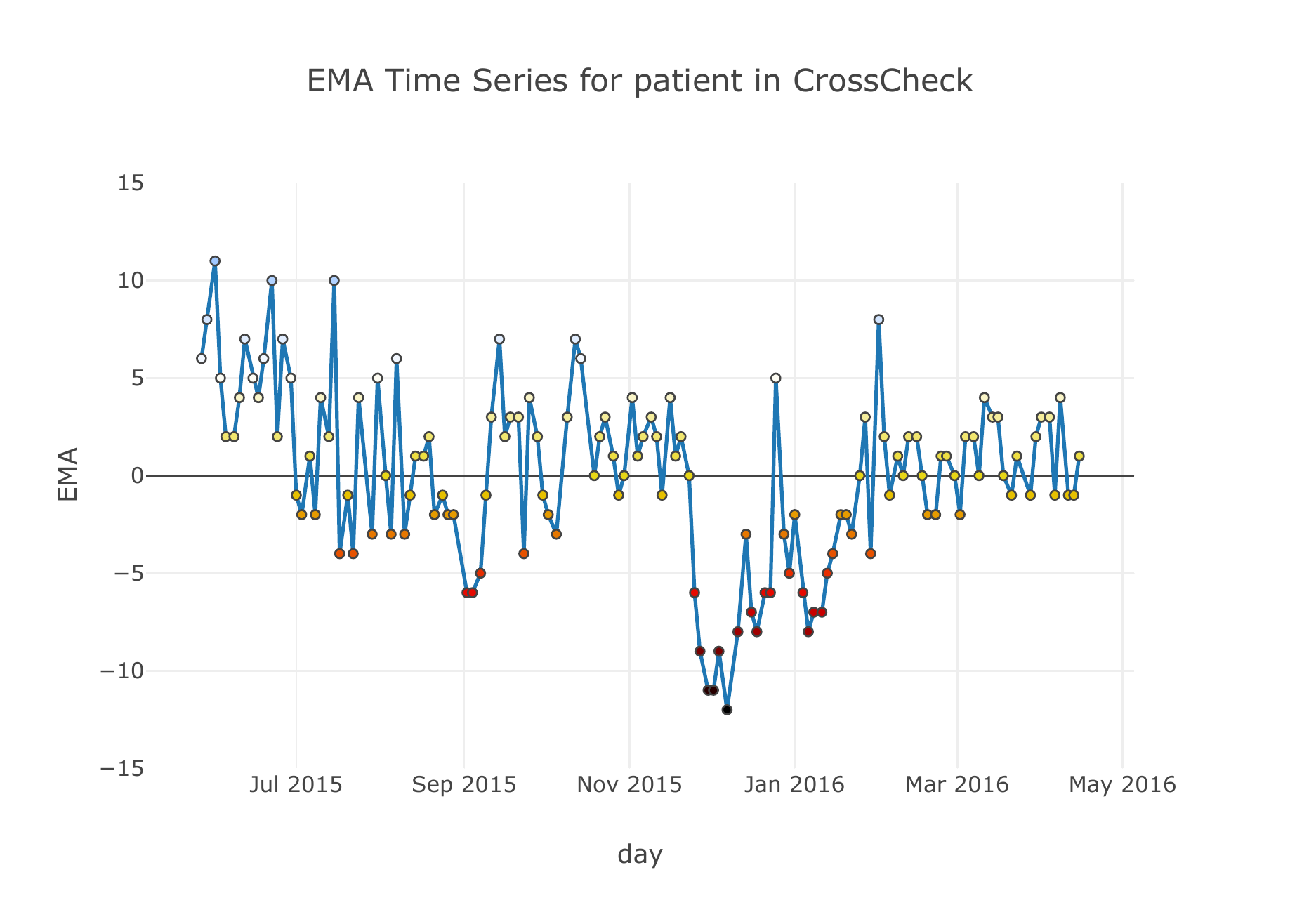}
  \caption{Here we show the trajectory of symptoms report for one user in the study. Starting from November 2015, there is a noticeable \textit{decrease} in the reported symptoms. Subsequently, the patient's symptoms gradually improve over time. Our focus lies in identifying the clinical interventions preceding this \ textit {decreases} in symptoms}
  \label{fig:patient3}
\end{figure}

Figure \ref{fig:patient3} shows the temporal fluctuations in behavioral data exhibited by patients. This graph illustrates the progress of a patient with SSDs who demonstrates high adherence to the treatment. Notably, following a period of relatively stable symptoms from July 2015 to November 2015, the patient reports a decrease in symptoms. In our study, we define a "decrease" as a period where, after experiencing positive symptom levels for some time, there is a subsequent diminishing of symptoms. It is crucial to differentiate a decrease from the clinical event known as a "relapse." A relapse encompasses various events such as psychiatric hospitalization, a significant increase in psychiatric care, the presence of suicidal ideation with clinical relevance, self-injury, or violent behavior resulting in harm to oneself or others \cite{Adler2020,buck2019relationships}. Relapse can be identified through assessments or electronic medical records, whereas decreases are self-reported reductions in symptoms. We hypothesize that digital phenotyping can be utilized to detect and analyze these decreases in symptoms.

Beyond benchmarks in predictive models, which have been the traditional focus of research in this field, we aim to shift our attention toward actionable models that can make a meaningful impact on clinical practice. The importance of behavioral feature analysis has been widely recognized in clinical practice \cite{tonekaboni2019clinicians}. In this study, we employ counterfactual explanations \cite{verma2020counterfactual}, a local interpretability method \cite{stiglic2020interpretability}, as a tool in IDSS for analyzing individual patient predictive models. Finally, we generate the feature importance of the clinical alert by evaluating the impact of \textit{decreases} in the clinical output. 

The clinical relevance of this research is to present a machine learning system as a potential tool for early intervention in patients. This system changes the conventional clinical practice towards a broader support system for physicians in decision-making. Concretely, our contributions are 1) Integration of a machine learning system in healthcare composed of prediction, detection, and explanation; 2) Using counterfactuals to inform which and when to intervene with relevant features in temporal settings with digital phenotyping. 

The prediction component of our machine learning system focuses on leveraging digital phenotyping data to develop models that can accurately forecast future symptom states for individual patients. These predictive models use historical patient data from various sources, such as self-reports and sensor data, to capture the temporal patterns and trends in patients symptoms. By training these models on a large dataset, we can identify the factors that contribute to symptom changes and make reliable predictions about future symptom trajectories.

In addition to prediction, the detection component of our system is designed to identify clinically relevant events, specifically decreases in symptoms, which may indicate a need for intervention. We employ change point detection algorithms to analyze the temporal patterns in symptom data and identify significant shifts or deviations from the expected behavior. This enables us to detect periods where patients experience a decrease in symptoms after a period of relatively stable or desired symptom states. By promptly detecting these symptom decreases, clinicians can intervene and provide appropriate support to prevent further health deterioration.

The interpretability component of our machine learning system, lies in \textit{explanations}, which utilizes counterfactuals research. Counterfactual explanations provide insights into the causal relationships between features and outcomes by generating hypothetical scenarios in which a specific feature is modified while keeping other variables constant. By applying counterfactual explanations to our predictive models, we can identify the features that are most influential in driving symptom changes and determine when and which interventions may be effective. This interpretability component empowers clinicians to understand the underlying mechanisms behind the predictions and make informed decisions regarding patient care.

By integrating prediction, detection, and explanation components into a unified clinical system, we aim to provide clinicians with actionable insights and decision support. This system has the potential to transform conventional clinical practice by incorporating data-driven approaches and enabling proactive interventions for patients with schizophrenia. By leveraging digital phenotyping data and state-of-the-art machine learning techniques, we can enhance the understanding and management of symptoms in real-world settings, leading to improved patient outcomes and personalized care.
\section{Related Work}
Digital Phenotyping aims to extract relevant clinical features from sensing data \cite{Mohr2017}. In this line of work, a stability index measurement was proposed by  \cite{He-Yueya} as an interpretable and clinically relevant feature for intervention. \cite{Tseng2020} presented a system that leverages human rhythms as features for symptom prediction using multi-task learning. They demonstrate the potential of using human rhythms for early detection and intervention without burdening patients or clinicians. The paper evaluates linear and non-linear models, highlighting the trade-off between prediction accuracy and interpretability.

The CrossCheck Project \cite{Wang2016, Ben-Zeev2017} conducted a clinical trial aimed at developing sensing, inference, and analysis techniques for detecting changes, relapse, prediction, and early intervention in patients with schizophrenia based on observational behavioral precursors. The project's clinical relevance lies in automatically alerting clinicians promptly to prevent or reduce the severity of relapse in patients with mid-symptoms of severe mental disorders. The CrossCheck symptom prediction system utilizes passive sensing and self-report data from phones to track schizophrenia patients' symptoms.  \cite{Wang2020b, Wang2020} utilized principal components as behavioral patterns and found that applying them reduces feature dimensionality and generates more useful features for prediction.  \cite{Adler2020} developed an encoder-decoder neural network-based anomaly detection model using passive sensing data to predict behavioral anomalies indicative of early warning signs for psychotic relapse. They conducted a post hoc analysis using clinical notes to interpret the detected anomalies within the context of severe mental disorders. While insightful, these approaches rely on clinical information to generate interpretations, making it challenging to adapt in real-world monitoring scenarios.

\cite{stachl2020predicting} explored the prediction of individuals' personality dimensions using smartphone behavioral data. The study investigated the predictive power of various classes of behavioral information, including communication, social behavior, music consumption, app usage, mobility, overall phone activity, and day- and night-time activity. However, the paper also acknowledges the potential privacy implications and dangers associated with the widespread collection and modeling of smartphone behavioral data and there is no actionable information in the study.

Beyond the CrossCheck project, we identified three lines of work related to the explanation of clinical time series. \cite{tonekaboni2020went} focused on explaining model predictions by identifying the relevant observations over time. They developed a generative model to capture the time series dynamics and determined feature importance by quantifying the shift in the predictive distribution over time. \cite{hardt2020explaining} used gradient methods and a dynamical system to capture predicted risk increases in clinical time series. \cite{crabbe2021explaining} explored computer vision methods to transform time series into images and applied perturbation-based detection methods to identify salient features. These three papers evaluated their tools using classical clinical data, such as intensive care unit data, to simulate real hospital monitoring scenarios.

In our paper, we employ the cumulative sum (CUSUM) algorithm for detecting decreases in self-reports and weekly predictions. The CUSUM algorithm is a sequential analysis technique originally introduced by \cite{page1954continuous} and has since found applications in various domains, including finance, industrial monitoring, public health monitoring, and clinical indicator analysis \cite{wohl1977cusum, sibanda2007cusum, suman2018control}.

Counterfactual explanations aim to provide insights into the decision-making process of predictive models by identifying the minimal set of changes required to alter the model's output. They allow exploration of hypothetical scenarios and answer questions such as "What if?" For a given input $\mathbf{x}$ with a model output of $y$, counterfactual explanations provide information on how changing $\mathbf{x}$ to $\mathbf{x}_{cf}$ would affect the model's output.

The Counterfactual Explanations (CFE) algorithm \cite{verma2020counterfactual} is a local interpretability method that enables specific queries and recourse for predictive models \cite{molnar2020interpretable}. By following the formulations and implementations in \cite{wachter2017counterfactual} and \cite{mothilal2020explaining}, we search for counterfactuals that possess key interpretability properties.

By using the CFE algorithm or similar approaches, researchers and practitioners can search for counterfactual explanations that possess important interpretability properties. These properties could include attributes like proximity to the original input, feasibility, and understandability, among others. By providing information on how changing the input would affect the model's output, counterfactual explanations enhance our understanding of predictive models and enable us to gain insights into their decision-making process.
\section{Methods}

We simulate a realistic setting of continuous clinical monitoring using digital phenotyping, following the clinical workflow presented in \cite{Wang2018}. The system generates reports on adherence and trends, enabling clinicians to proactively reach out to patients when the system predicts an increased risk. To further enhance interpretability, we propose an additional step of local interpretability that identifies the location of potentially relevant clinical events and the features that should be intervened upon.

\begin{figure*}[ht]
  \centering
  \includegraphics[scale=0.4]{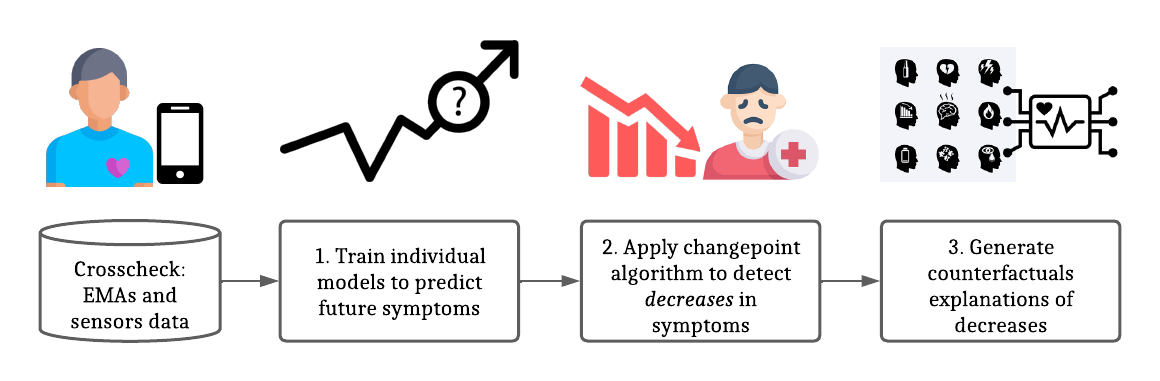}
  \caption{The machine learning system consists of prediction, detection, and explanation components. This pipeline is applied to individualized models using more than one month of digital phenotyping data}
  \label{fig:pipeline}
\end{figure*} 

\paragraph{System overview.} In the depicted setting shown in Figure \ref{fig:pipeline}, we expect to observe a patient's data for a duration longer than one month. Subsequently, we generate symptom predictions based on the available historical data (Step 1 in Figure \ref{fig:pipeline}). For each weekly prediction, we employ a change point algorithm that identifies potential decreases in symptoms by analyzing the time series comprising previous self-reports and predicted self-reports (Step 2 in Figure \ref{fig:pipeline}). If a clinical alert in the form of a change point is detected, we apply counterfactual explanations to the weekly predictions (Step 3 in Figure \ref{fig:pipeline}). This workflow can be viewed as a predictive clinical tool that supports decision-making in a continuous monitoring setting.

\paragraph{Problem Statement.}
Let $X \in \mathbb{R}^{d \times n}$ denote a multi-variate time series data where $d$ is the number of features with $n$ observations over time and $\mathbf{y} \in\mathbb{R}^{n}$ a univariate time series of size $n$. At each time step $t = 1, 2, ..., n$ a real value vector, $\mathbf{x}_t = (x_t^1,x_t^2,...,x_t^d)^\intercal \in \mathbb{R}^d$ and  the scalar $y_t \in \mathbb{R}$ is provided.  We consider the problem of explaining an alert in a time segment $i:j$ where $ i < n < j$ include both historical values (where $t\leq n$) and \textit{predictive} (where $t>n$) of the output $y_t$, that is, an alert generated from a temporal dynamic of the vector $(y_{i},...,y_{n},...,y_{j})$. We are interested in actionable information from the set of features $X$ to inform physicians about possible explanations of the provided alert.

\paragraph{Dataset.}

The CrossCheck dataset was collected from a  psychiatric hospital in New York. The patients were 18 years old or older, met DSM criteria for schizophrenia, schizoaffective disorder, or psychosis, and had a psychiatric hospitalization, daytime psychiatric hospitalization, outpatient crisis management, or short-term psychiatric hospital emergency room visits within 12 months before the study entry \cite{Wang2016}. In the dataset, 61 out of 75 participants completed the full year-long study. 

Ecological Momentary Assessments (EMAs) or self-reports are clinically validated questions that capture various dynamic dimensions of mental health and functioning in individuals with schizophrenia. The study employs EMAs comprising 10 questions, as presented in Table \ref{tab:ema_questions}. A higher score on the positive questions indicates better outcomes, while higher scores on the negative item questions suggest worse outcomes. The questions are formulated as concise one-sentence queries, and respondents choose from multiple-choice answers ranging from 0 to 3 \cite{Wang2016}.

The second kind of data is passive data from raw sensors collected without the active intervention of patients. The behavioral sensing features are composed of sensors about activity, speech, conversation, calls and SMS, sleep location, phone and app usage, and ambient environment.  The dataset is composed of aggregated values of sensors for each day at different time epochs: morning (6 am to 12 pm), afternoon (12 pm to 6 pm), evening (6 pm to 12 am), and night (12 am to 6 am). A more detailed description of this dataset can be found in \cite{Wang2016,Wang2020}.

\begin{table}[]
\centering
\begin{tabular}{|c|l|}
\hline
Category                  & \multicolumn{1}{c|}{Question}              \\ \hline
\multirow{5}{*}{Positive} & \, 1. Have you been feeling CALM?             \\
                          & \, 2. Have you been SOCIAL?                   \\
                          & \, 3. Have you been SLEEPING well?            \\
                          & \, 4. Have you been HOPEFUL about the future? \\
                          & \, 5. Have you been able to THINK clearly?    \\ \hline
\multirow{5}{*}{Negative} & \begin{tabular}[c]{@{}l@{}} \, 6. Have you been worried about people trying to \\   \> \,  \, HARM you?\end{tabular}  \\
                          & \, 7. Have you been bothered by VOICES?       \\
                          & \, 8. Have you been feeling STRESSED?         \\
                          & \begin{tabular}[c]{@{}l@{}}\, 9.  Have you been SEEING THINGS other people \\ \> \,     \, can not see?\end{tabular} \\
                          & 10. Have you been DEPRESSED?               \\ \hline
\end{tabular}
\caption{Questionnaire related to indicators of mental health used in the CrossCheck project. Options: 0—not at all; 1—a little; 2—moderately; 3—extremely}
\label{tab:ema_questions}
\end{table}

\paragraph{Preprocessing EMAs.}
In our context, we define a \textit{block} as a segment of data that is continuous in time. To construct a block, we consider two parameters: the minimum size of a block and the maximum distance between each data point.
For the minimum size of a block, we use 60 days, which corresponds to 15 data points. This ensures that we have enough data to build predictive models and make reliable predictions.

Regarding the maximum distance between each data point, we set a limit of 15 days (equivalent to 6 data points) based on previous preprocessing techniques described in Wang et al. \cite{Wang2016}. This means that we consider a gap of up to 15 days between self-report data points within a block.

Furthermore, previous studies \cite{Adler2020, Wang2020} have shown that there are behavioral patterns and signs of fluctuations close to the 30-day mark, which could be indicative of an impending relapse. Therefore, it is more suitable to think of the \textit{weekly} continuous monitoring as a monitoring period consisting of 3 data points within the block range. By defining blocks in this way, we can effectively analyze and interpret the temporal patterns and fluctuations in the data for predictive modeling and monitoring purposes.

In our filtering approach, we employ a variance filter on the EMAs time series. This helps us identify self-reports that do not exhibit dynamic behavior or fluctuations over time. Since our research is specifically interested in analyzing fluctuations in symptoms, a patient whose symptoms remain constant throughout would not provide relevant insights for our research question.
By applying this filter, we aim to include patients whose self-reports reflect meaningful variations in their symptoms, thereby ensuring that our machine-learning and counterfactual models capture the dynamic nature of mental health conditions and provide more accurate predictions and insights.

\paragraph{Predictive Modeling.}

In the first part of the framework in Figure \ref{fig:pipeline} we use individual forecast EMA sum scores using past values of each 10 past EMAs or sensors. An individual model is a fully personalized model, which uses data only from the subject to train the model. Several studies in the CrossCheck project have shown that the individual models outperform population-level ones \cite{Wang2016, He-Yueya}. Formally, we consider a multivariate time series $X \in \mathbb{R}^{d \times n}$ and $Y \in\mathbb{R}^{n}$ a univariate time series, where $d$ is the number of features with $n$ observations over time. We are interested in the multi-step ahead point forecasting problem that involves producing an estimate of the $H$ future values $y_{n+1},y_{n+2},...,y_{n+H}$, where $H>1$ is the forecast horizon using a data drive prediction model composed by a predictive function $f$ and a set of features $X$.

\paragraph{Hyper-parameter tuning.}
We search hyperparameters for several predictors. Motivated by the result of \cite{Tseng2020} where they found that nonlinear models have higher prediction accuracy than linear models. We explore the trade-off between linear and nonlinear models as in \cite{zihni2020opening} \cite{stachl2020predicting}. In our case, we compare linear and ensemble models to forecast weekly symptoms using digital phenotyping. Specifically, we compare linear models such as Lasso and Elastic Net and ensemble methods such as Random Forest and Gradient Boosting. After obtaining the optimal hyperparameters, we select the best model and use it in the rest of the system.

For each block, the data were ordered and split into training and test sets with a corresponding $80\%$ and $20\%$. This approach allowed us to test for optimal model settings while ensuring a strict separation of training and test data, especially in the data-dependent case. Then the models were tuned using 5-fold time series forward cross-validation. Table \ref{tab:hyper} provides a summary of the tuned hyperparameters for each model. We used a grid search approach for the tuning of hyperparameters in all models \cite{pedregosa2011scikit}.

\paragraph{Model training.} We used the best-tuned predictors in the simulated clinical workflow. In this part, we test performance using the time series forward (rolling origin) Cross-Validation approach \cite{bergmeir2012use}. In this case, for each block, the minimum date of the test set is always bigger than the maximum date of the training set. In this order, the size of the training set always increases and the size of the test set is always the size of weekly predictions. This is coherent with a real scenario of monitoring and the time-series properties. For testing, we use the Mean Absolute Error (MAE) that corresponds to the expected value of the absolute error loss: $\text{MAE}(y, \hat{y}) = \frac{1}{H} \sum_{k=n+1}^{n+H}  |y_k - \hat{y_k}|$. 

\paragraph{Change point Detection.}
Here we describe the change point detection applied to the symptoms time series, as depicted in the second part of the framework shown in Figure \ref{fig:pipeline}. A change point refers to an abrupt shift in a time series, indicating a significant change in the underlying dynamics \cite{basseville1993detection}. The goal of change point detection is to identify the times within a given time horizon when such changes occur.

The strength of the CUSUM approach \cite{page1954continuous} lies in its simplicity and visual clarity. The algorithm operates online, making it suitable for early detection. We begin by defining a segment of the study, denoted as $\mathbf{y}{i:j} = (y_i,y{i+1},...,y_{j-1}, y_j)^\intercal $, which represents a portion of the time series. A reference point, typically chosen as the middle point of the segment, divides it into two subsegments. We calculate the means of each subsegment, denoted as $\mu_{a}$ and $\mu_{b}$, respectively. We then compute the mean difference, $\mu = \frac{\mu_{a}- \mu_{b}}{2}$. Next, we apply the CUSUM function to find the argument that maximizes the cumulative sum of the difference between each data point in the segment of study and $\mu$. Formally, we aim to solve the following optimization problem:

\begin{equation}
\begin{aligned}
\arg \max_{k} c(y[i:j]) = \sum_{k=i+1}^N \Vert y[k] - \mu \Vert \text{ for } N = i+1, ..., j.
\end{aligned}
\end{equation}
The algorithm used in our approach estimates the means before and after a potential change point. It iteratively searches for the change point by maximizing the cumulative sum value until either a stable change point is found in the segment or the maximum number of iterations is reached \cite{truong2020selective}.

To adapt this algorithm to the continuous monitoring setting, we employ a sliding window approach. For each week, we perform CUSUM searches within a window of interest. This window consists of a historical window, which includes the previous 12 data points, a scan window, which includes the next 6 data points, and a step size of 1 data point. Following the search process described above, we evaluate the potential change point using a Gaussian distribution as the underlying model. We apply the log-likelihood ratio test to determine if there is a statistically significant change in the mean of the time series. The null hypothesis assumes no change in the mean, while the alternative hypothesis suggests a change point with two distinct means \cite{kats}. In the continuous monitoring setting, we collect all the potential change points detected within each week and select the most recent change point identified across the all-time series. If a change point is detected within a 2-week window (consisting of 1 week of historical data and 1 week of predictions), we trigger a clinical alert. This approach enables the timely detection of significant changes in the time series, allowing for proactive interventions when necessary.

\paragraph{Counterfactual Explanations.}
The third part of the framework in Figure \ref{fig:pipeline} focuses on explaining the alerts generated by the change point algorithm using the predictive model. To achieve this, we employ the counterfactual explanations algorithm (CFE) \cite{verma2020counterfactual}, which aims to identify the minimal set of changes necessary to alter the model's output.
CFE operates by searching for hypothetical scenarios, asking questions like: "Given that the model $f$ produces output $y$ for input $\mathbf{x}$, what would be the output if $\mathbf{x}$ were changed to $\mathbf{x}_{cf}$?" In other words, it determines the changes needed in $\mathbf{x}$ to achieve a desired change in the model's output. CFE serves as a local interpretability method, enabling specific queries about the predictive model \cite{molnar2020interpretable}.
We follow the formulation and implementations presented in \cite{wachter2017counterfactual} and \cite{mothilal2020explaining}, seeking counterfactuals with certain properties. These properties include being \textit{proximal} to highlight the local decision logic of the predictor, \textit{sparse} to emphasize a limited set of features, \textit{diverse} to showcase different ways of achieving the same outcome, and \textit{feasible}, meaning the changes in a counterfactual example should be within the possible range of each feature.

Consider a $d$-dimensional vector $\mathbf{x}$ representing a specific instance at a given time $t$. We have a machine learning model $f$ that provides predictions $f(\mathbf{x})$, where the predicted values fall within a certain domain. Within this domain, some subsets correspond to desired outcomes, while others correspond to undesired outcomes. The goal of counterfactual explanation is to find a counterfactual instance $\mathbf{x}{cf}$ that satisfies the following conditions: (1) $f(\mathbf{x}{cf})$ falls into the desired outcome subset, (2) $\mathbf{x}{cf}$ is close to $\mathbf{x}$, and (3) $\mathbf{x}{cf}$ is both feasible and plausible.

To obtain a counterfactual explanation, we formulate an optimization problem:
\begin{equation}
\begin{aligned}
\text{arg} \min_{\mathbf{x}{cf}} & \quad \text{dist}(\mathbf{x},\mathbf{x}{cf})\\ 
& \quad \textrm{s.t.} \quad & f(\mathbf{x}{cf}) \text{ falls within the desired outcome range}\\ 
&&& \mathbf{x}{cf} \in P\\
&&& \mathbf{x}{cf} \in F(\mathbf{x})
\end{aligned}
\end{equation}
Here, $\text{dist}$ represents a distance function that measures the proximity between $\mathbf{x}$ and $\mathbf{x}{cf}$. The counterfactual explanation operates within the domain space $\mathcal{X} \subset \mathbb{R}^{d}$ and is subject to plausibility constraints denoted by $P$ and feasibility constraints denoted by $F(\mathbf{x})$. These constraints ensure that the counterfactual instance is both plausible and feasible.

In our specific case, we employ counterfactual explanations using a forecasting model that predicts the EMA sum score. This means that $f(\mathbf{x})$ represents the predicted EMA sum score based on the features in $\mathbf{x}$. By generating counterfactual instances that satisfy the conditions outlined above, we can provide explanations for instances where the predicted EMA sum score falls into the undesired outcome subset. These counterfactual explanations help us understand the changes necessary in the input features to achieve a desired outcome, making them valuable for decision-making and intervention planning.

\section{Results}

\begin{figure*}[ht]
  \centering
  \includegraphics[scale=0.45]{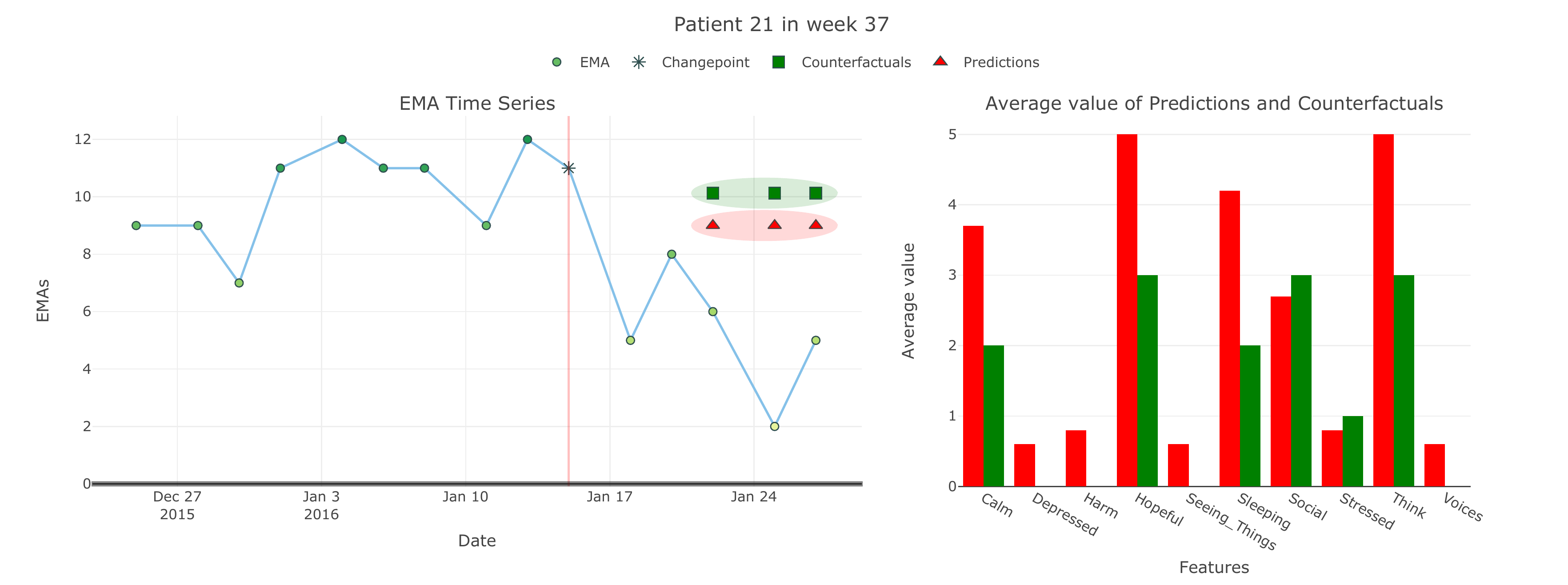}
\caption{Results of continuous symptom monitoring and counterfactual explanation.
Historical data (left) shows symptom values of 12 to 7 before January 17.
In under a week, decreasing symptoms were detected and counterfactuals generated (right).
The system identifies symptom changes, determines modifications needed, and helps prevent adverse health events.}
\label{fig:patient}
\end{figure*}

In Figure \ref{fig:patient} (left), a complete execution of our simulated clinical workflow for a specific patient is depicted. The system initiates by monitoring relevant outcomes in the form of time series. Predictions are generated (depicted as red triangles). Subsequently, the change point algorithm is applied to the merged historical and weekly predictive time series. If a change point is detected, the CFE algorithm generates a hypothetical situation where the patient exhibits improved symptoms (represented by green squares). Clinicians can then intervene in the features that differ between the predictions and the counterfactuals.

In Figure \ref{fig:patient} (right), the alert generated highlights which features change and the magnitude of their changes. This information enables our system to identify the location of an alert, the features involved, and the extent of changes required to achieve the desired outcome.

After preprocessing the dataset, we obtained a total of 44 patients and 51 blocks. The minimum block size is 26 data points (equivalent to approximately two months and one week), while the maximum block size is 165 data points. On average, each block contains 91 data points.

\paragraph{Definitions.} \textit{Validity} refers to the fraction of examples that are actually counterfactuals, meaning they result in a different outcome than the original input. \textit{Proximity} represents the average distance between a counterfactual example and the original input, calculated feature-wise. Proximity is evaluated by taking the mean of the distances for a set of examples. \textit{Sparsity} captures the number of features that are different between the original input and a generated counterfactual. It is defined as the number of changes between the two inputs. \textit{Diversity} can be evaluated by measuring the feature-wise distances between each pair of counterfactual examples. The mean distance between pairs represents the diversity of the set. Separate diversity metrics are computed for categorical and continuous features. \textit{Redundancy} refers to the duplicated or highly similar information across multiple counterfactual explanations. High redundancy means the method is proposing the same or nearly identical explanations repeatedly, indicating limited diversity. \\
\begin{table}[]{
\begin{tabular}{rrrrrr}
\hline
\textbf{features} &
  \textbf{method} &
  \multicolumn{1}{l}{\textbf{validity}} &
  \multicolumn{1}{l}{\textbf{redundancy}} &
  \multicolumn{1}{l}{\textbf{sparsity}} &
  \multicolumn{1}{l}{\textbf{proximity}} \\ \hline
\multirow{3}{*}{\textbf{EMAs}}    & \textbf{genetic} & 0.9(0.10) & \textbf{0.19(0.16)} & 0.75(0.16)          & \textbf{0.45(0.46)} \\
                                  & \textbf{kdtree}  & 1.0(0.0) & 0.15(0.18)          & 0.79(0.2)           & 0.34(0.43)          \\
                                  & \textbf{random}  & 1.0(0.0) & 0.15(0.08)          & \textbf{0.82(0.04)} & 0.18(0.12)          \\ \hline
\multirow{3}{*}{\textbf{Sensors}} & \textbf{genetic} & 0.85(0.11) & \textbf{0.67(0.31)} & 0.22(0.14)          & \textbf{0.53(0.57)} \\
                                  & \textbf{kdtree}  & 0.91(0.09) & 0.67(0.32)          & 0.21(0.14)          & 0.52(0.6)           \\
                                  & \textbf{random}  & 1.0(0.0) & 0.52(0.23)          & \textbf{0.4(0.11)}  & 0.01(0.01)          \\ \hline
\multirow{3}{*}{\textbf{\begin{tabular}[c]{@{}r@{}}Sensors\\ +EMAs\end{tabular}}} &
  \textbf{genetic} &
  0.9(0.13) &
  0.6(0.33) &
  0.25(0.13) &
  \textbf{1.72(11.41)} \\
                                  & \textbf{kdtree}  & 0.87(0.11) & \textbf{0.62(0.32)} & 0.24(0.13)          & 1.71(11.4)          \\
                                  & \textbf{random}  & 1.0(0.0) & 0.45(0.25)          & \textbf{0.44(0.1)}  & 0.01(0.01)          \\ \hline
\end{tabular}

}
\caption{Evaluation of counterfactual explanations methods}
\label{tab:my-tabledd}
\end{table}

The research results presented in Table \ref{tab:my-tabledd} compare the performance of different counterfactual explanation methods across three feature sets: EMAs, Sensors, and Sensors+EMAs. The evaluation is based on four key metrics: validity, redundancy, sparsity, and proximity. The counterfactual explanation methods employed are genetic, kdtree, and random.

The results show that all methods consistently achieved high validity scores, indicating their effectiveness in producing meaningful counterfactual explanations. The genetic method consistently outperforms the other methods in terms of redundancy for both the EMAs and Sensors feature sets. This suggests that the genetic method generates a more diverse set of counterfactuals by exploring different regions of the feature space. The random method consistently exhibited the highest sparsity across all feature sets, implying that it introduces a larger number of changes compared to the other methods. Lower proximity scores indicate that the counterfactuals are closer to the original input. Interestingly, there is no clear pattern among the methods in terms of proximity scores, as the values vary across feature sets and methods.

Evaluating counterfactual generation methods requires consideration of validity in relation to redundancy, sparsity, proximity, and diversity. A nuanced understanding of performance across metrics can determine the optimal method for developing meaningful counterfactual explanations in a given context and support application to diverse datasets.

\subsection{Predictive performance evaluation}

\begin{figure}[]
  \includegraphics[scale=0.6]{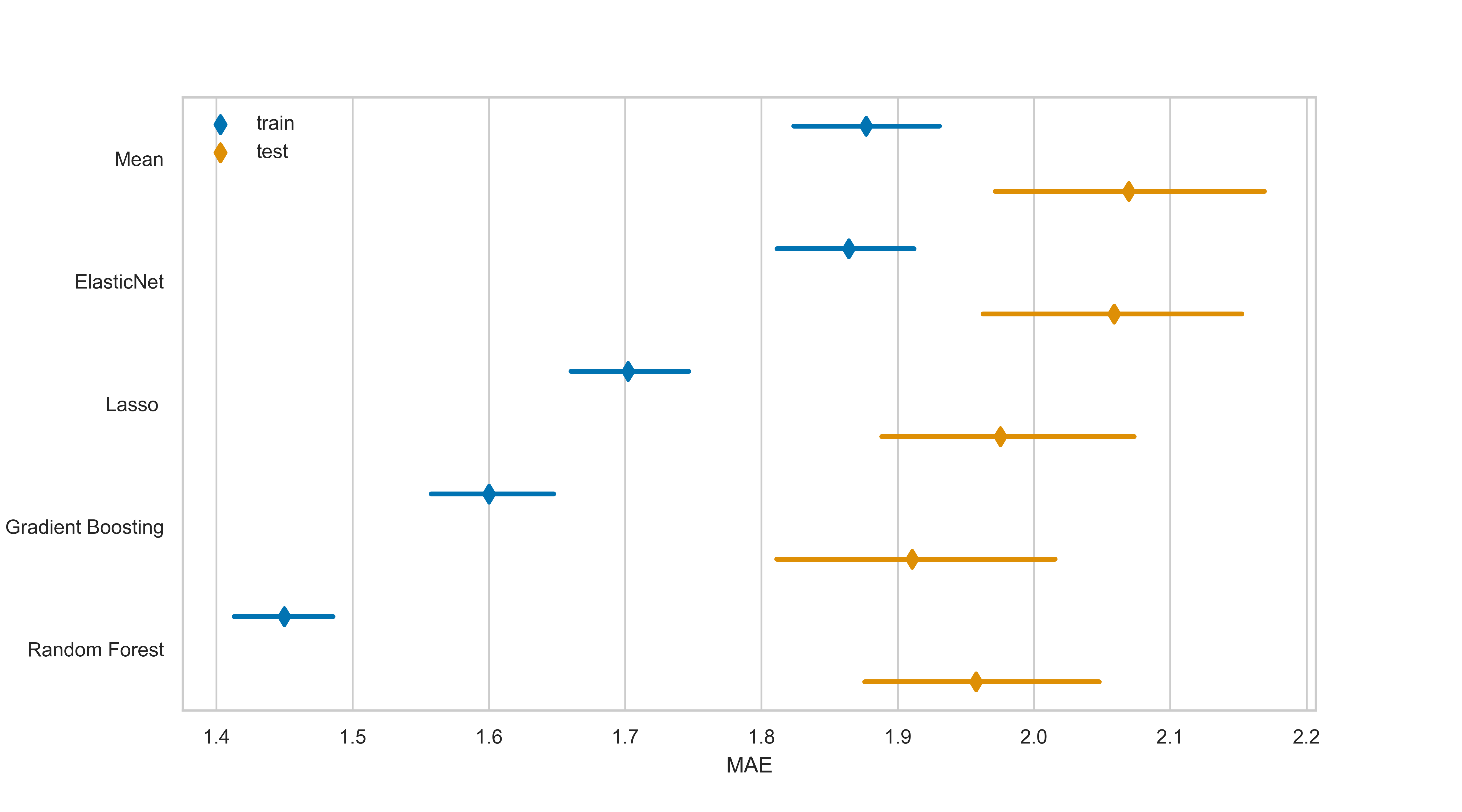}
  \caption{Comparison of Train and Test Error with MAE for Different Models.}
  \label{fig:models}
\end{figure}

In our simulated setting, we used the information from the three previous self-reports to forecast the EMA sum score. Individual predictions were initiated with a minimum of 12 data points and increased weekly. As depicted in Figure \ref{fig:models}, all predictors demonstrated comparable performance. These models were trained using 12 data points based on the week of predictions, and the training data consisted of all EMA questions. The results show that there is minimal variation among the models in this setup. The error bars represent 95\% bootstrapped confidence intervals. Specifically, the MAE values in the test set were as follows: mean baseline - 2.070, Elastic Net - 2.059, Lasso - 1.975, Random Forest - 1.958, and GBRT - 1.911. Although the GBRT model exhibited the highest performance, the difference in performance among the models was marginal. It is worth noting that the nonlinear models consistently outperformed the linear prediction performance on average. Therefore, we selected the GBRT model for all subsequent predictions in this paper.

After conducting the cross-validation process, we successfully determined the optimal hyperparameters for each predictor. The Lasso model yielded an alpha value of 0.74. For the Elastic Net model, we obtained an alpha of 10 and an L1 ratio of 0.14. The Random Forest model was configured with a maximum tree depth of 5, consisting of 10 trees. We set the minimum number of samples required to split an internal node to 2, and the minimum number of samples required at a leaf node to 3. Lastly, the Gradient Boosting model utilized a Huber loss function, a learning rate of 0.05, 10 boosting stages, a maximum depth of estimators set to 3, and minimum numbers of samples required at a leaf node and to split an internal node set to 5.

\begin{figure}
  \includegraphics[scale=0.6]{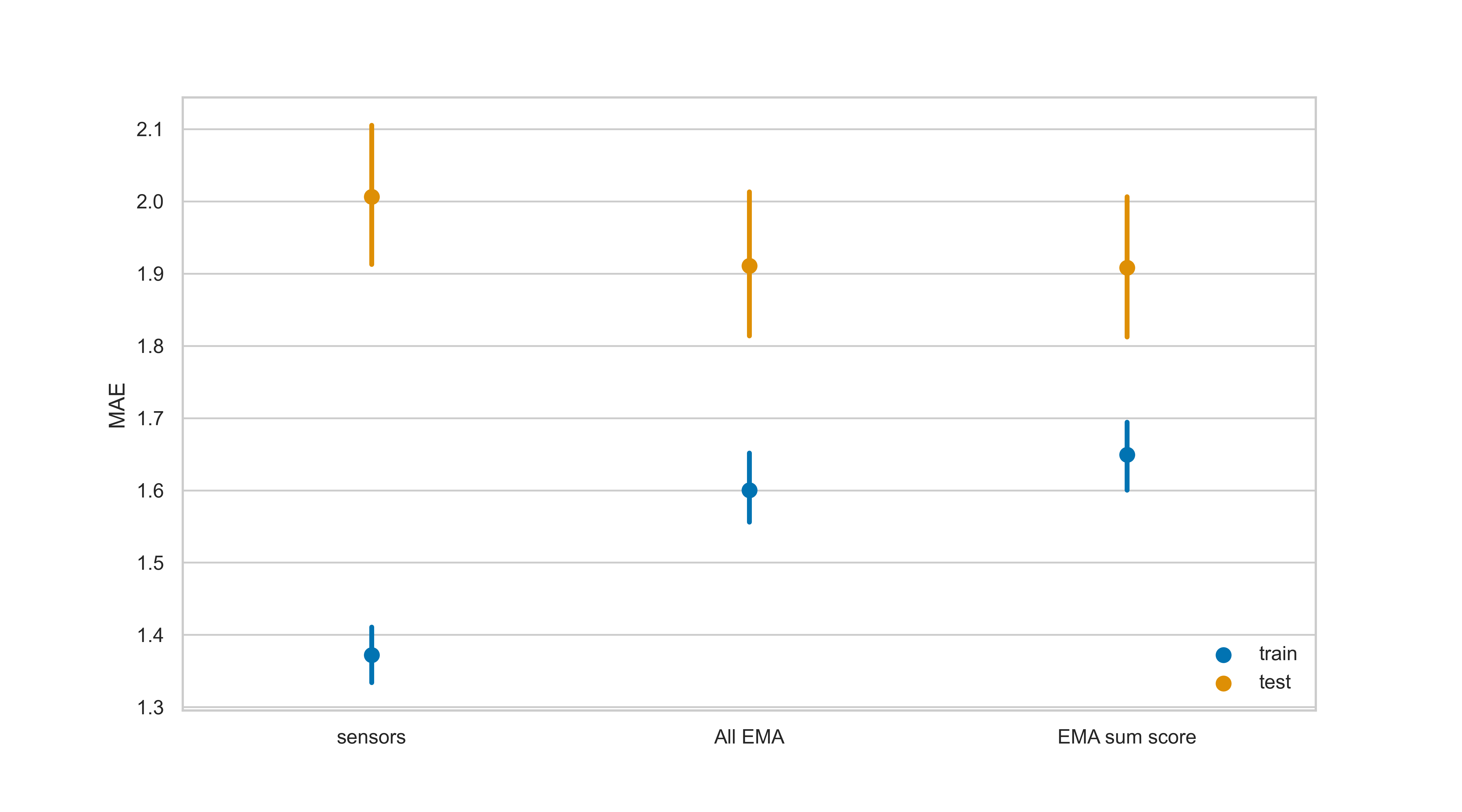}
  \caption{Features performance for the GBRT-Huber model. There are no big differences between different features. This opens the possibility of using raw sensors as forecasting features. The error bars indicate $95\%$ bootstrapped confidence intervals.}
  \label{fig:features}
\end{figure}

Figure \ref{fig:features} presents a comparison of feature performance for the GBRT-Huber model. The features evaluated include the historical EMA sum score, all EMA responses, and sensor data. The error bars in the figure represent bootstrapped confidence intervals at a 95\% level.

As shown in Figure \ref{fig:features}, there are no significant differences in performance among the different features. This finding suggests that it is feasible to use raw sensor data as forecasting features, opening the possibility of leveraging sensor information for behavior prediction.

By utilizing the GBRT model and conducting weekly predictions, we obtained the following mean MAE values: 1.911 for all past EMAs, 1.908 for the past EMA sum score, and 2.006 for the sensor data. Although all past EMAs and the past EMA sum score demonstrate slightly better performance, we select all past EMAs as the feature set for our final model. This choice is based on its clearer clinical interpretation and relevance to the forecasting task.

These results are encouraging, as they suggest that raw sensor data can be leveraged for forecasting  behavior and support the application of CFE techniques to sensor data.

\subsection{Change points Detection of predicted symptoms}
\autoref{tab:my-table} presents an evaluation of different changepoint algorithms based on their recall and F1-score, with a 5-day delay.

The Baseline Zero algorithm achieves a recall of 0.53 and the highest F1-score of 0.66. It strikes a good balance between true positives and false positives, leading to a high F1-score. However, its relatively lower recall suggests that it may miss some true positive change points. This algorithm serves as a baseline for comparison.

The CUSUM algorithm exhibits a higher recall of 0.73 compared to Baseline Zero. However, its F1-score is lower at 0.52. This indicates that CUSUM detects more true positive changepoints but also produces more false positives, resulting in a lower overall F1-score.

The CUSUM algorithm with sliding window achieves a recall of 0.61 and an F1-score of 0.56. Although its recall is slightly lower than the original CUSUM algorithm, it manages to improve the F1-score. This suggests a better balance between true positives and false positives compared to the standard CUSUM algorithm.

The Bayesian Online algorithm demonstrates the highest recall of 0.77 among all the evaluated algorithms. However, its F1-score is 0.58, indicating that it also produces a significant number of false positives. Despite this, it may be suitable for applications where high recall is crucial.

The Robust Detection algorithm achieves a recall of 0.64 and an F1-score of 0.51. It exhibits a relatively higher recall compared to the CUSUM algorithm but has a lower F1-score. This implies that it detects more true positive changepoints but also generates more false positives, resulting in a lower overall F1-score.
\begin{table}[]
\centering
\begin{tabular}{|l|l|l|}
\hline
\textbf{\begin{tabular}[c]{@{}l@{}}Changepoint \\ Algorithm\end{tabular}} & \textbf{Recall} & \textbf{F1-Score} \\ \hline
Baseline Zero    & 0.53          & \textbf{0.66} \\ \hline
CUSUM            & 0.73          & 0.52          \\ \hline
\begin{tabular}[c]{@{}l@{}}CUSUM \\ sliding window\end{tabular}           & 0.61            & 0.56              \\ \hline
Bayesian Online  & \textbf{0.77} & 0.58          \\ \hline
Robust Detection & 0.64          & 0.51          \\ \hline
\end{tabular}
\caption{Evaluation of Changepoint Algorithms using recall and F1 score using 5 days of delay. The best algorithm in terms of recall is the CUSUM with a sliding window. The baseline zero algorithm is the best in terms of F1 score, this could be explained by many false-negative reports. This implies a risk in terms of alarm fatigue.}
\label{tab:my-table}
\end{table}

In summary, the evaluation of these changepoint algorithms reveals trade-offs between recall and F1-score. The CUSUM algorithm with a sliding window strikes a balance between the two metrics. The Baseline Zero algorithm achieves the highest F1-score but has a lower recall, indicating a risk of false-negative reports. On the other hand, the Bayesian Online algorithm has the highest recall but also a notable number of false positives. The choice of algorithm will depend on the specific application and the relative importance of recall and precision in the clinical context.

\begin{figure}
  \includegraphics[scale=0.7]{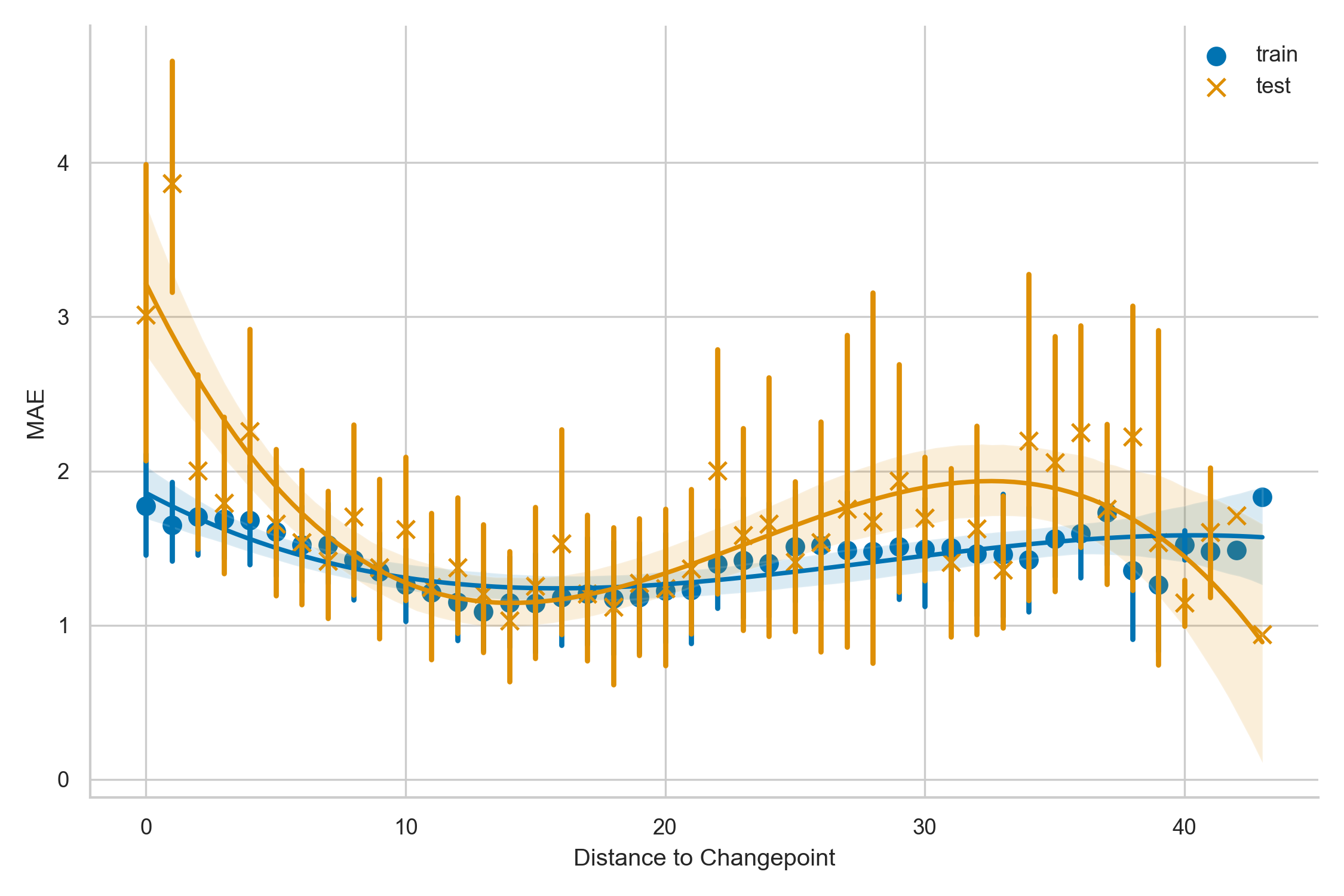}
  \caption{Relationship between distance to a change point and forecasting error. The plot illustrates that as data points approach the change point, the errors in forecasting increase. This observation is likely attributed to the distribution shift occurring in the training data.}
  \label{fig:distribution_shift}
\end{figure}

We have identified a total of 33 change points distributed across 22 blocks. However, during our analysis, we encountered an unexpected observation regarding the influence of the distance to change points on the algorithm's error. To measure this distance, we consider the number of weeks between a data point and a change point. Figure \ref{fig:distribution_shift} visually demonstrates that as a data point gets closer to a change point, the forecasting error increases. This phenomenon exemplifies a classic case of distribution shift, which has the potential to impact machine learning systems. Notably, recent research has explored the concept of distribution shift to developing alert systems \cite{tonekaboni2020went}. In our future work, we aim to focus on developing prediction models and CFE algorithms that are robust to distribution shifts \cite{rawal2020algorithmic}. Understanding and addressing the challenges posed by distribution shifts in prediction models and CFE algorithms are crucial for the robustness and reliability of our system. By further investigating and mitigating the effects of distribution shifts, we can enhance the performance and effectiveness of our algorithms in real-world scenarios.

\subsection{Counterfactual Explanations of Predicted Symptoms}

\begin{figure}
  \includegraphics[scale=0.6]{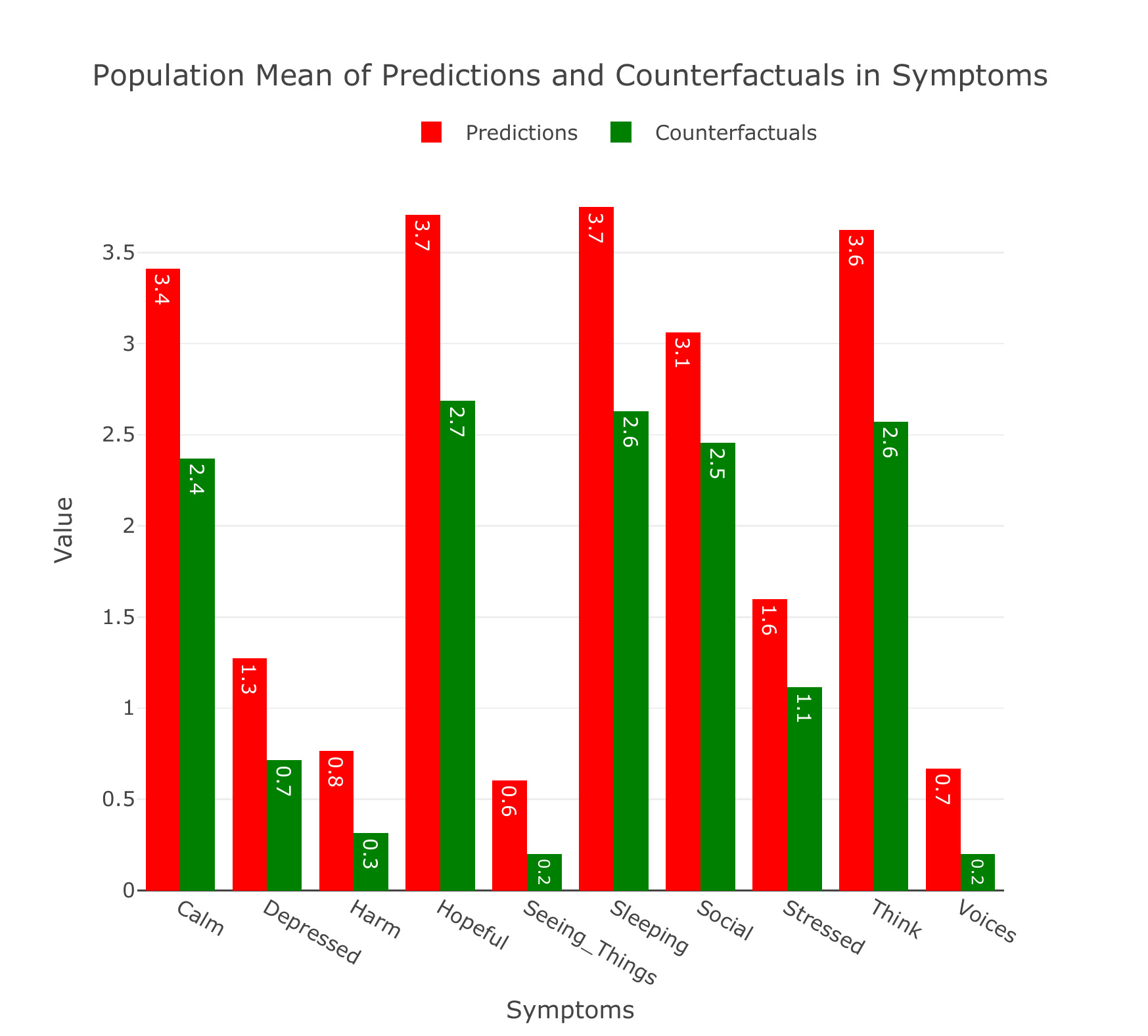}
  \caption{Population values for each EMA question by predictions and counterfactuals.}
  \label{fig:population}
\end{figure}

Figure \ref{fig:population} illustrates the transitions between predictions and counterfactual results, showing the pre-explanation values of each feature and the necessary changes for all patients. Clinicians can use these transitions to identify specific areas requiring attention and potential interventions for each patient.

\begin{figure}
  \includegraphics[scale=0.7]{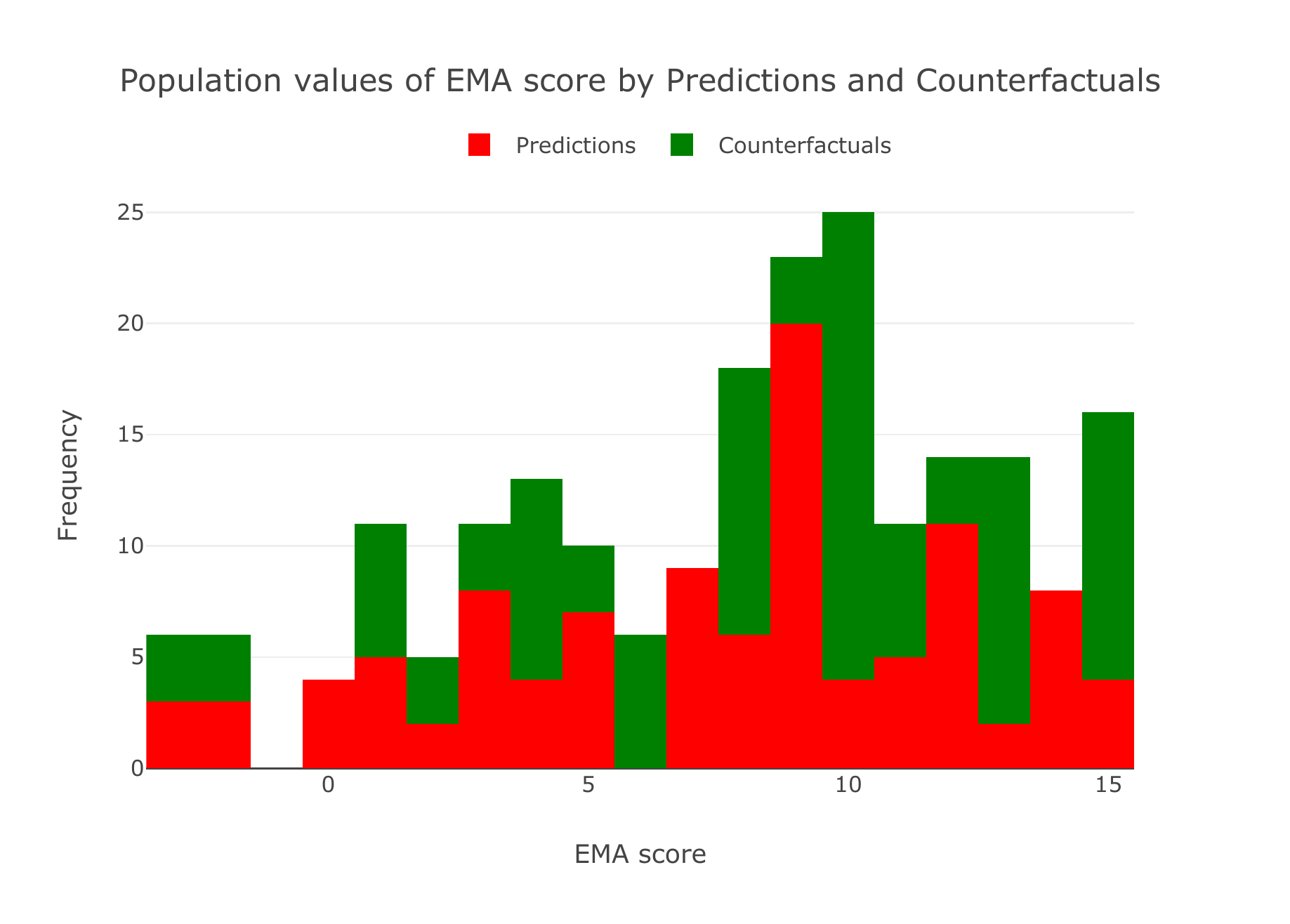}
  \caption{Histogram of values of EMA sum score by predictions and counterfactuals. We can see that detection of decreases still in high symptoms.}
  \label{fig:pred_cfe}
\end{figure}

We generated a total of 33 counterfactual explanations during our simulated clinical workflow. Figure \ref{fig:pred_cfe} depicts the histogram of EMA sum score values for predictions and counterfactuals. The observed decreases in symptoms generally fall within desirable ranges, which may be attributed to the scarcity of patients with low symptoms in the CrossCheck project and the system's limited ability to detect abrupt changes. Notably, a separate study \cite{Adler2020} identified four relapse patients using clinical labeled data, not EMAs.

In conclusion, our counterfactual explanations provide valuable insights into predicted symptom changes. Although the detected decreases mostly occur within higher symptom ranges, future work should focus on establishing a ground truth to evaluate the accuracy of alerts. These enhancements will improve the reliability and effectiveness of our counterfactual explanation system, supporting clinical decision-making and patient care.

\section{Conclusions}

In conclusion, our study investigated various aspects of a simulated clinical workflow, including counterfactual explanations, predictive performance evaluation, and change point detection of predicted symptoms in schizophrenia patients.

\textit{Counterfactual Explanation Methods}: The evaluation of different counterfactual explanation methods revealed that all methods achieved high validity scores, indicating their effectiveness in generating meaningful counterfactual explanations. The genetic method consistently outperformed the other methods in terms of redundancy, suggesting that it generated a more diverse set of counterfactuals. The random method exhibited the highest sparsity. 

\textit{Predictive Performance Evaluation}: The comparison of different predictive models showed minimal variation in performance, with marginal differences in MAE values. Nonlinear models consistently outperformed linear models, with the GBRT model exhibiting the highest performance. The choice of the GBRT model for subsequent predictions was based on its slightly better performance and the clinical interpretation of its features.

\textit{Change Point Detection}: The evaluation of different change point detection algorithms showed trade-offs between recall and F1-score. The CUSUM algorithm with a sliding window achieved a balance between the two metrics, while the Baseline Zero algorithm had the highest F1-score but lower recall. The Bayesian Online algorithm had the highest recall but also more false positives. The choice of algorithm depends on the specific application and the importance of recall and precision in the clinical context.

Overall, our study provides insights into the performance and capabilities of a simulated clinical workflow, highlighting the potential of using counterfactual explanations, interpretability, predictive models, and change point detection for improving patient care and decision-making. Further research can build upon these findings and explore additional aspects of the workflow to enhance its effectiveness and applicability in real-world healthcare settings.

This study has also limitations. First, all models were trained on a small sample of individuals with SSDs. Additionally, assessments of symptoms were reported only on selected days. The small dataset may have contributed to reduced predictive performance, further work should apply this pipeline with clinical datasets with more granular levels (minutes or seconds) and large sample datasets. Additionally, the change point algorithm does not have a ground truth to test the accuracy of alerts. The CFE algorithm assumes that the features in the prediction algorithm are symmetrically related to the outcome. It is important to construct robust prediction models and evaluate the change point and CFE algorithms before deployment in real-world settings.

\section{Future directions}

In our study, we primarily relied on EMAs and sensor data for predictive modeling and counterfactual analysis. However, there is a wealth of other data sources that can be incorporated to improve the accuracy and robustness of the models. For instance, integrating electronic health records (EHRs), genetic data, social determinants of health, and additional clinical variables can provide a more comprehensive understanding of a patient's health status and enable more accurate predictions. Future work should explore the integration of diverse data sources to create a holistic view of patient's health profiles and enable personalized interventions.

To fully understand the potential impact of our approaches, it is essential to evaluate their clinical utility and effectiveness. Conducting studies to assess the impact of using predictive models and counterfactual explanations in clinical decision-making can provide valuable insights into their value in improving patient outcomes, reducing healthcare costs, and enhancing the quality of care. Future work should focus on conducting rigorous clinical trials and outcome evaluations to quantify the benefits and limitations of the proposed approaches.

As with any technology in healthcare, there are ethical considerations associated with the use of predictive models and counterfactual explanations. Future research should address ethical challenges related to data privacy, consent, algorithmic bias, and the responsible deployment of these technologies in clinical practice. Incorporating ethical frameworks and involving stakeholders, including patients, clinicians, and policymakers, in the design and evaluation process can help mitigate ethical concerns and ensure the responsible use of these tools.

 Developing techniques to provide meaningful and transparent explanations for these models would greatly enhance their usability and acceptance in critical domains such as healthcare. Moreover, exploring the integration of counterfactual explanations with other interpretable techniques, such as rule-based models or symbolic reasoning, could provide a more holistic and comprehensive understanding of the model's decision-making process.

\section*{Acknowledgment} 
Research supported by the National Science Foundation (NSF) under Grant No. 1918839. We thank Armando Solar-Lezama and the computer-assisted programming group at CSAIL MIT for fruitful discussions.

\bibliographystyle{ACM-Reference-Format} 
\bibliography{references}
\newpage
\appendix

\section{Appendix}

\paragraph{Hyperparameter space and description for predictive performance evaluation.} After conducting the cross-validation process, we successfully determined the optimal hyperparameters for each predictor. The Lasso model yielded an alpha value of 0.74. For the Elastic Net model, we obtained an alpha of 10 and an L1 ratio of 0.14. The Random Forest model was configured with a maximum tree depth of 5, consisting of 10 trees. We set the minimum number of samples required to split an internal node to 2, and the minimum number of samples required at a leaf node to 3. Lastly, the Gradient Boosting model utilized a Huber loss function, a learning rate of 0.05, 10 boosting stages, a maximum depth of estimators set to 3, and minimum numbers of samples required at a leaf node and to split an internal node set to 5.

\begin{table}[ht]
\centering
\begin{tabular}{|c|c|c|}
\hline
\textbf{Predictor} &
  \textbf{Hyperparameter} &
  \textbf{Values} \\ \hline
Lasso &
  \begin{tabular}[c]{@{}c@{}}Alpha (Constant that\\ multiplies the L1 term)\end{tabular} &
  \begin{tabular}[c]{@{}c@{}} $\{ \frac{k}{100} : 0 < k <100,$ \\  
  $ k \in \mathbb{N} \}$
  \end{tabular} \\ \hline
\multirow{2}{*}{\begin{tabular}[c]{@{}c@{}}Elastic\\ Net\end{tabular}} &
  \begin{tabular}[c]{@{}c@{}}L1 ratio (The ElasticNet \\ mixing parameter)\end{tabular} &
  \begin{tabular}[c]{@{}c@{}}$\{ \frac{k}{100} : 0 < k <100,$ \\  
  $ k \in \mathbb{N} \}$\end{tabular} \\ \cline{2-3} 
 &
  \begin{tabular}[c]{@{}c@{}}Constant that multiplies \\ the penalty terms\end{tabular} &
  \begin{tabular}[c]{@{}c@{}}0.00001 ,0.0001, 0.001, \\ 0.01, 0.1, 0, 0.5, 1, 10, 100\end{tabular} \\ \hline
\multirow{5}{*}{\begin{tabular}[c]{@{}c@{}}Random \\ Forest\end{tabular}} &
  \begin{tabular}[c]{@{}c@{}}Whether bootstrap \\ samples are used \\ when building trees\end{tabular} &
  True, False \\ \cline{2-3} 
 &
  \begin{tabular}[c]{@{}c@{}}The maximum depth\\  of the tree\end{tabular} &
  5, 10, 20, 50, 100 \\ \cline{2-3} 
 &
  \begin{tabular}[c]{@{}c@{}}The number of trees\\  in the forest.\end{tabular} &
  \begin{tabular}[c]{@{}c@{}}5, 10, 50, 100, \\ 500, 1000\end{tabular} \\ \cline{2-3} 
 &
  \begin{tabular}[c]{@{}c@{}}The minimum number\\  of samples required to \\ split an internal node\end{tabular} &
  1, 2, 5 \\ \cline{2-3} 
 &
  \begin{tabular}[c]{@{}c@{}}The minimum number\\  of samples required \\ to be at a leaf node\end{tabular} &
  1, 3, 5 \\ \hline
\multirow{6}{*}{\begin{tabular}[c]{@{}c@{}}Gradient \\ Boosting\end{tabular}} &
  Loss function &
  Huber, Squared Error \\ \cline{2-3} 
 &
  Learning rate &
  \begin{tabular}[c]{@{}c@{}}0.001, 0.01, 0.05, \\ 0.1, 0.2, 1\end{tabular} \\ \cline{2-3} 
 &
  \begin{tabular}[c]{@{}c@{}}The number of\\  boosting stages\end{tabular} &
  \begin{tabular}[c]{@{}c@{}}5, 10, 50, 100, \\ 500, 1000\end{tabular} \\ \cline{2-3} 
 &
  \begin{tabular}[c]{@{}c@{}}Maximum depth \\ of the individual \\ regression estimators\end{tabular} &
  2, 3, 5, 10 \\ \cline{2-3} 
 &
  \begin{tabular}[c]{@{}c@{}}The minimum number \\ of samples required \\ to be at a leaf node\end{tabular} &
  1, 5, 10, 20 \\ \cline{2-3} 
 &
  \begin{tabular}[c]{@{}c@{}}The minimum number \\ of samples required \\ to split an internal node\end{tabular} &
  2, 5, 10, 20 \\ \hline
\end{tabular}
\caption{Model hyperparameters and values. These models were tuned before the simulated clinical workflow using time series 5-fold cross-validation using 80\% of each model. The 20\% of each model was used to evaluate.}
\label{tab:hyper}
\end{table}
\end{document}